\documentclass[a4paper]{article}

\usepackage[pdftex]{graphicx}
\usepackage{float}
\usepackage{ascmac}
\usepackage{theorem}
\usepackage{longtable}
\usepackage{bm}

\usepackage[fleqn]{amsmath}
\usepackage{latexsym}
\usepackage{newtxtext}
\usepackage[varg]{newtxmath}

\usepackage[rm]{caption}

\newcommand{\setdef}[2]{\left\{\left.{#1}\;\right|\;{#2}\right\}}

\title{Block-Segmentation Vectors for Arousal Prediction using Semi-supervised Learning}
\author{Yuki ODAKA\thanks{Department of Computer and Network Engineering, Graduate School of Informatics and Engineering, The University of Electro-Communications} \and Ken KANEIWA\fontsize{14pt}{14pt}\selectfont $^*$}
\date{}

\begin{document}
\maketitle

\begin{abstract}
To handle emotional expressions in computer applications, Russell's circum- plex model has been useful for representing emotions according to valence and arousal. In SentiWordNet, the level of valence is automatically assigned to a large number of synsets (groups of synonyms in WordNet) using semi-supervised learning. However, when assigning the level of arousal, the existing method proposed for SentiWordNet reduces the accuracy of sentiment prediction. In this paper, we propose a block-segmentation vector for predicting the arousal levels of many synsets from a small number of labeled words using semi-supervised learning. We analyze the distribution of arousal and non-arousal words in a corpus of sentences by comparing it with the distribution of valence words. We address the problem that arousal level prediction fails when arousal and non-arousal words are mixed together in some sentences. To capture the features of such arousal and non-arousal words, we generate word vectors based on inverted indexes by block IDs, where the corpus is divided into blocks in the ﬂow of sentences. In the evaluation experiment, we show that the results of arousal prediction with the block-segmentation vectors outperform the results of the previous method in SentiWordNet.

\end{abstract}


\maketitle

\section{Introduction}

Human emotions have been investigated for a long time in order to analyze product reviews and tweets in opinion mining \cite{picasso2019technical}\cite{grgic2021application}\cite{xu2018emo2vec}\cite{biddle2020leveraging}. In particular, the main approach to determining a writer's emotions from sentences is to predict the valence, which is whether a word or sentence is positive or negative. Moreover, because it is diﬃcult to express various emotions with valence alone, diﬀerent emotion models have been proposed. The circular model \cite{russell1980circumplex} is one such model that can express emotions from multiple perspectives by adding arousal to valence. Another model of factor analysis deals with arousal and dominance in addition to valence \cite{osgood1957measurement}\cite{russell2003core}\cite{xiang2021affective}.



To determine the sentiment of a sentence, an emotion dictionary, in which sentiment levels are assigned to words, is needed. However, in practice, it is difficult to manually label the sentiment levels of a large number of words in documents. Therefore, the labels are manually assigned to only a few words and are used as training data for machine learning algorithms. However, supervised learning tends to reduce the accuracy of the sentiment prediction of unlabeled words because of the shortage of manually labeled training data. To solve this problem, we adopt semi-supervised learning to expand the number of labeled words by predicting the sentiment level of each word step by step from a small number of labeled words.


SentiWordNet \cite{esuli2006sentiwordnet}\cite{baccianella2010sentiwordnet} is the emotional dictionary in which the valence level of each synset (a group of synonyms) is added in the English semantic dictionary WordNet \cite{miller1995wordnet}. In this approach, the valence levels of many synsets are automatically scored by semi-supervised learning, in which a few words manually labeled from WordNet's deﬁnition corpus are used for training.


We consider that semi-supervised learning-based arousal prediction cannot achieve a high accuracy in the previous method proposed in SentiWordNet. Unlike words with similar valence, arousal words do not appear together in some sentences. This makes it difficult to predict arousal levels from the surrounding words because the distribution of similar words aﬀects the word vectors generated from sentences that are used as input vectors for machine learning algorithms.


In this paper, we propose a block-segmentation vector for predicting the arousal levels of many synsets from a small number of labeled words using semi-supervised learning. We analyze the distribution of arousal and non-arousal words in a corpus of sentences by comparing it with the distribution of valence words. To adapt to the difficulty of arousal prediction, we exploit a corpus of Wikipedia example sentences instead of the WordNet corpus of deﬁnition sentences and example sentences. The long sentences of the corpus enhance the representation of word features, which can take into account the ﬂow of sentences. For this purpose, we generate word vectors based on inverted indexes by block IDs, where the corpus is divided into many blocks. Furthermore, we deal with feature selections of the word vectors for arousal prediction based on the information gain ratio (IGR) and the Gini coefficient.


The remainder of this paper is organized as follows: Section 2 gives an overview of SentiWordNet. Section 3 describes our proposed method for arousal prediction, which consists of the block-segmentation word vectors and the semi-supervised learning model. In Section 4, we present an experimental evaluation of the block-segmentation word vectors for arousal prediction. Finally, in Section 5, we conclude this paper and discuss future work.

\section{SentiWordNet}

\subsection{Valence prediction in WordNet}

\begin{figure}[t]
	\begin{center}
		\includegraphics[width=10cm]{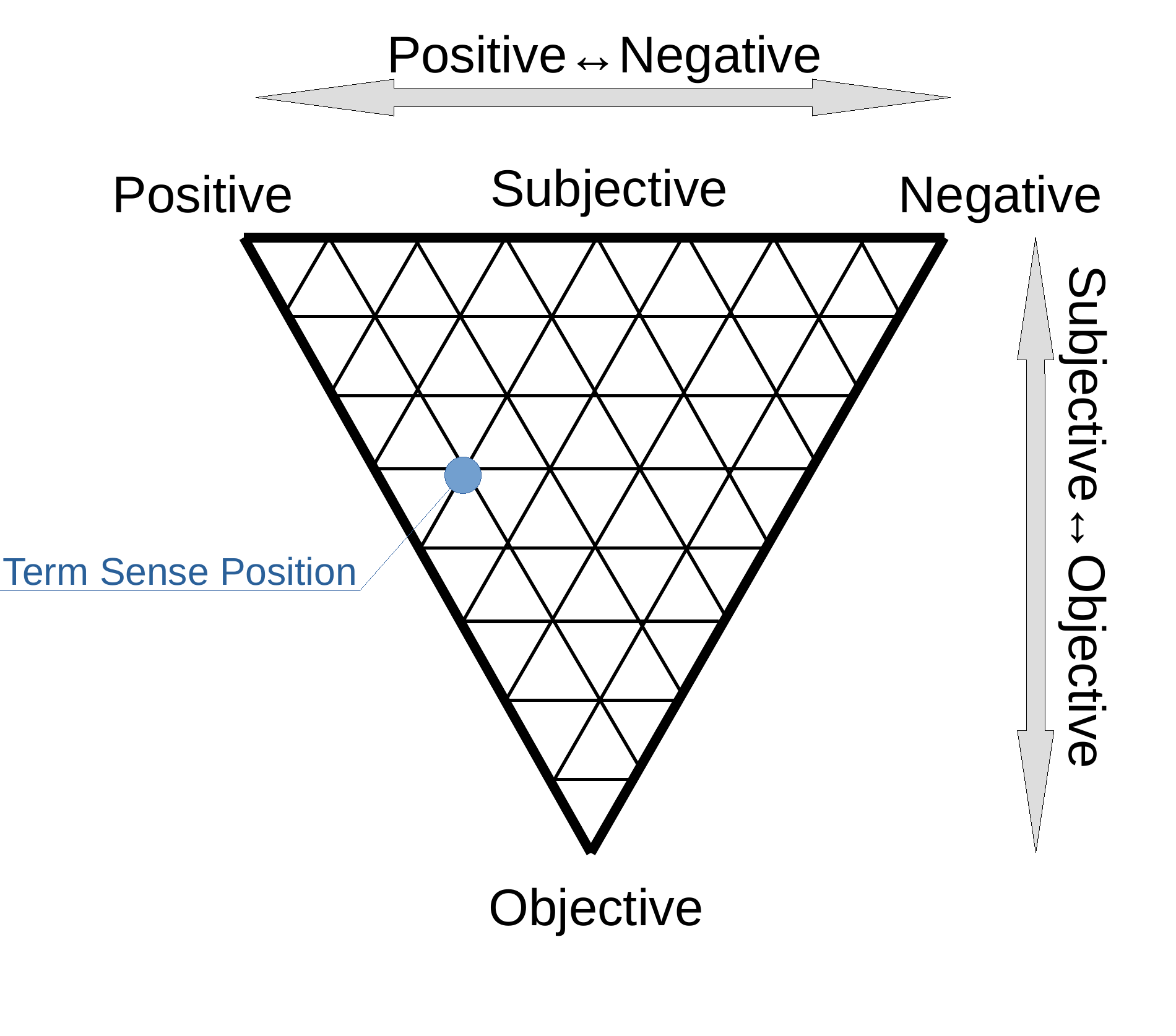}
		\caption{Model of SentiWordNet \cite{esuli2006sentiwordnet}} 
		\label{fig:SentiWordNetPosition}
	\end{center}
\end{figure}

\begin{figure}[t]
	\begin{center}
		\includegraphics[width=12cm]{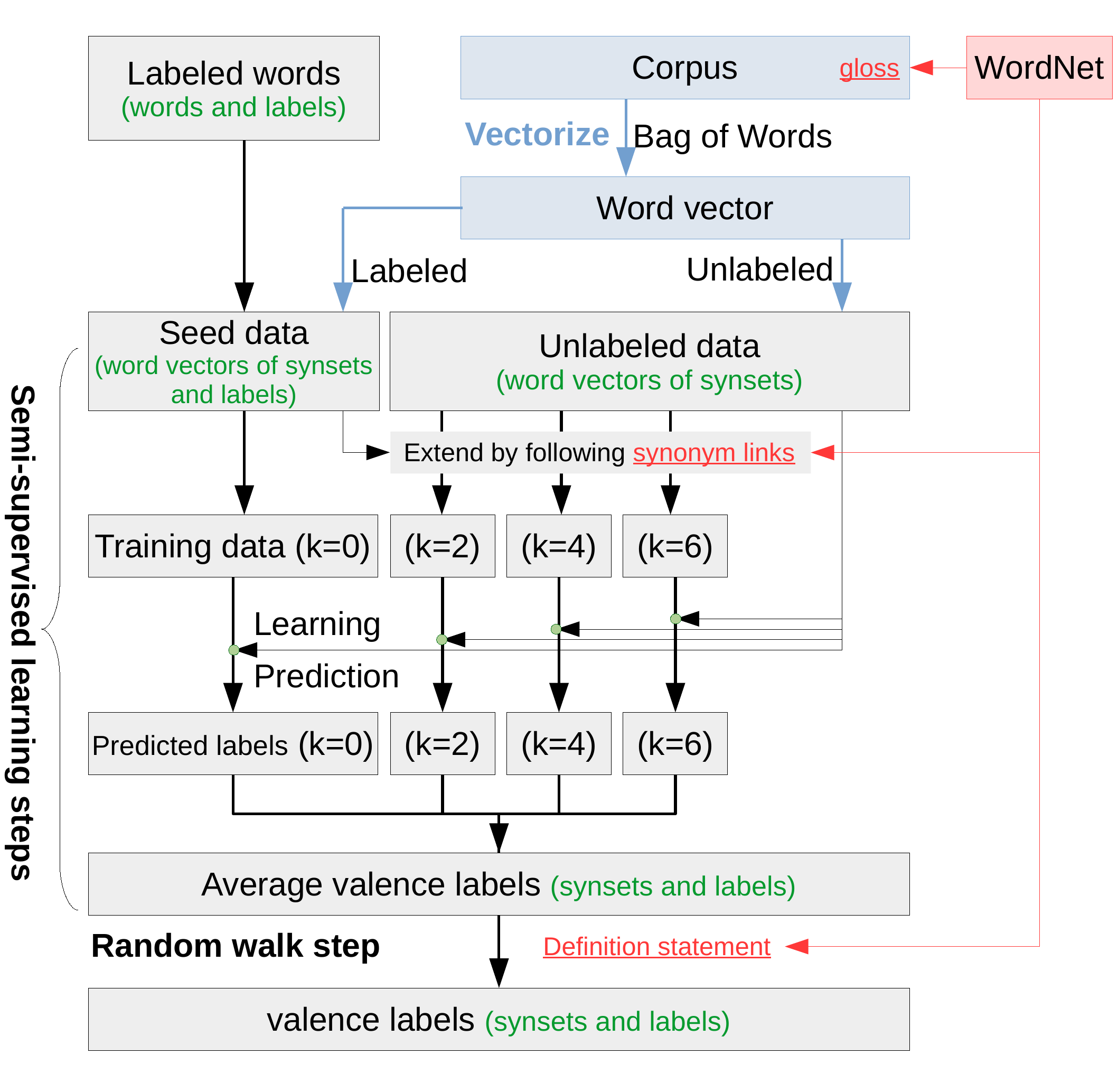}
		\caption{Flow of valence prediction} 
		\label{fig:flow}
	\end{center}
\end{figure}


WordNet is an English semantic dictionary that contains 117,659 synsets associated with synonyms. In the dictionary synonyms and concept hierarchies are represented by links between synsets. For each synset, there are deﬁnition sentences and example sentences, which are called glosses.



SentiWordNet \cite{esuli2006sentiwordnet} is an emotional dictionary that assigns two kinds of valence levels: Positive$\leftrightarrow$Not-Positive and Negative$\leftrightarrow$Not-Negative to each synset in Word- Net. The valence is based on a model consisting of the two axes: Positive$\leftrightarrow$Negative and Subjective$\leftrightarrow$Objective, as shown in Figure  \ref{fig:SentiWordNetPosition}. Using the predicted valence levels, we can determine that the more Not-Positive and Not-Negative the synset is, the more Objective it is. The process of valence prediction consists of (i) semi-supervised learning steps and (ii) a random walk step, as shown in Figure 2.

\subsubsection{Semi-supervised learning steps} 

The vector representation of a synset (called the word vector) is generated from its glosses (e.g., deﬁnition sentences) by the bag-of-words (BoW) method \cite{zhang2010understanding}. The following seven positive words $\textit{Pwords}$ and seven negative words $\textit{Nwords}$ are given as the source data for labeling some of the synsets in WordNet \cite{turney2003measuring}.
\begin{align}
\hspace{-16pt} \textit{Pwords} =& [\textrm{"good," "nice," "excellent," "positive," "fortunate," "correct," "superior"}] \nonumber \\
\hspace{-16pt} \textit{Nwords} =& [\textrm{"bad," "nasty," "poor," "negative," "unfortunate," "wrong," "inferior"}] \nonumber
\end{align}


The synsets in which these words are synonyms are used as seed data. A synset with a radius of $k$ can be reached in $k$ steps from the seed data by following WordNet links. 


In semi-supervised learning steps, synsets with a radius of $k$ are labeled by following synonym links from the seed-data. The valence levels of Positive$\leftrightarrow$Not-Positive are assigned to the synsets reached from the $\textit{Pwords}$ as positive examples and the $\textit{Nwords}$ as negative examples. These labeled synsets are used as training data in machine learning methods, and then the labels of the remaining synsets are predicted by the trained model. SentiWordNet uses two machine learning methods: the Rocchio algorithm and a support vector machine (SVM) , to perform binary classiﬁcation tasks for four radii $k \in \{0, 2, 4, 6\}$ where Positive is 1 and Not-Positive is 0. We predict a total of eight classes $c_{k,\textrm{Rocchio}}$ and $c_{k,\textrm{SVM}}$ using the trained eight models of the four binary classifications and two learning methods. Then, the valence level of each synset is calculated by the average of eight predicted classes $\sum_{k \in \{ 0,2,4,6 \}}\frac{c_{k,\textrm{Rocchio}} + c_{k,\textrm{SVM}}}{8}$. Moreover, the valence levels for Negative$\leftrightarrow$Not-Negative are predicted in the same way.

\subsubsection{Random walk step}


In SentiWordNet, the semi-supervised learning steps \cite{esuli2007random} are followed by a random walk step to improve the accuracy of label prediction. That is, we reflect the predicted labels of synsets included in the deﬁnition sentences when labeling the synsets of their headwords in the next process.


Let $G=\langle N,L \rangle$ be a graph where $N$ is a set of nodes and $L$ is a set of edges. Each $n_i \in N$ represents a synset. If a word in synset $n_j$ appears in the deﬁnition sentence of a headword in synset $n_i$, then there is a directed edge $(n_j, n_i) \in L$ from $n_j$ to $n_i$. Let $C(i) = \{n_j \in N \; | \; (n_j,n_i) \in L\}$, and let $a_1, \ldots, a_{|N|}$ be the labels of synsets $n_1 , \ldots , n_{|N|}$. Then, the random walk step is defined by the following update equation from $a_j$ to $a_i$.
\begin{align}
\label{exp:randomWalk}
a_i^{(k)} \leftarrow \frac{\alpha}{|C(i)|} \sum_{n_j \in C(i)}{a_j^{(k-1)}} + (1 - \alpha)e_i
\end{align}
where $a_i^{(k)}$ is the value of $a_i$ at the $k$th iteration, $\alpha$ is a control parameter with $0\le\alpha \le 1$, and $e_i$ is a constant with $\sum_{1 \le i \le |N|}{e_i} = 1$ and $e_i = a_i^{(0)}$.

\begin{figure}[t]
	\begin{center}
		\includegraphics[width=12cm]{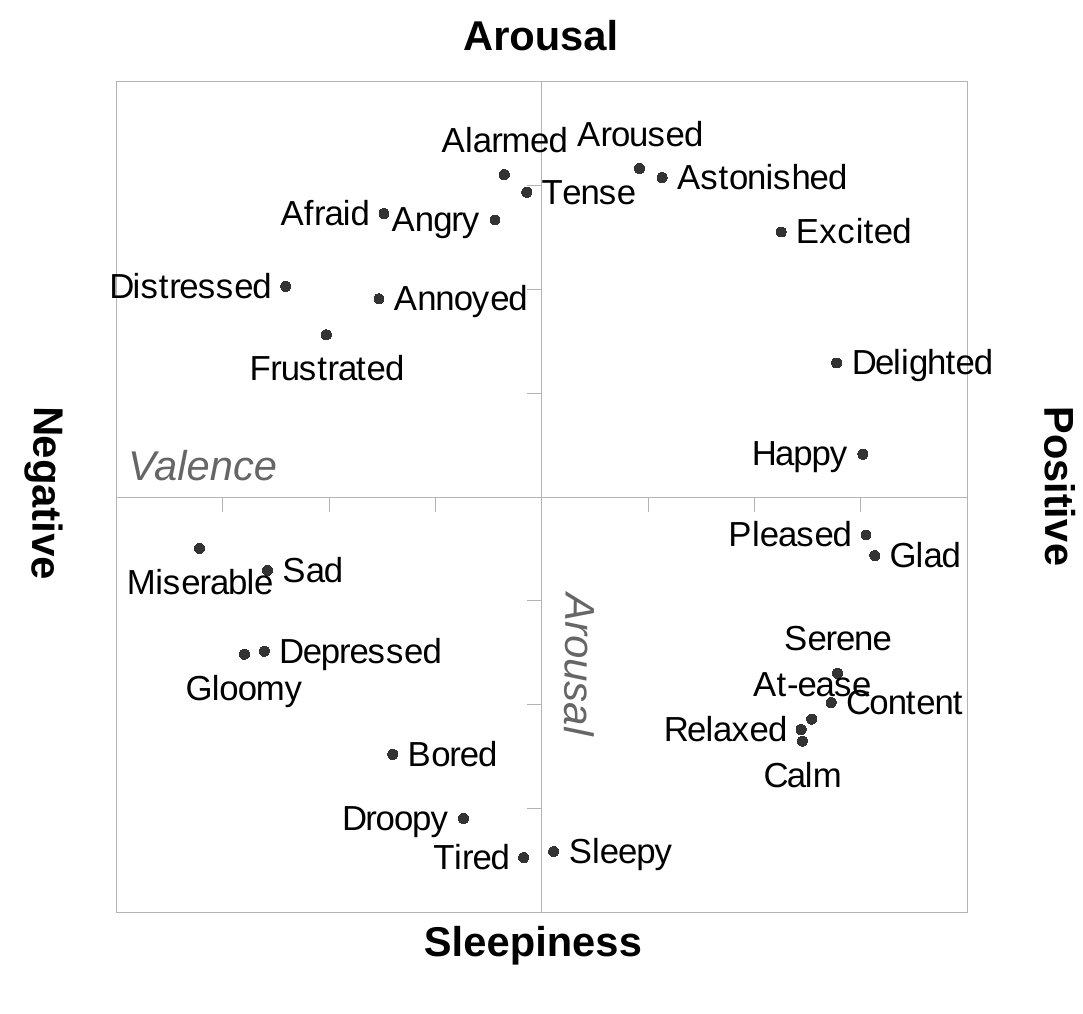}
		\caption{Russell's circumplex model \cite{russell1980circumplex}} 
		\label{fig:russell}
	\end{center}
\end{figure}

\begin{figure}[t]
	\begin{center}
		\includegraphics[width=8.5cm]{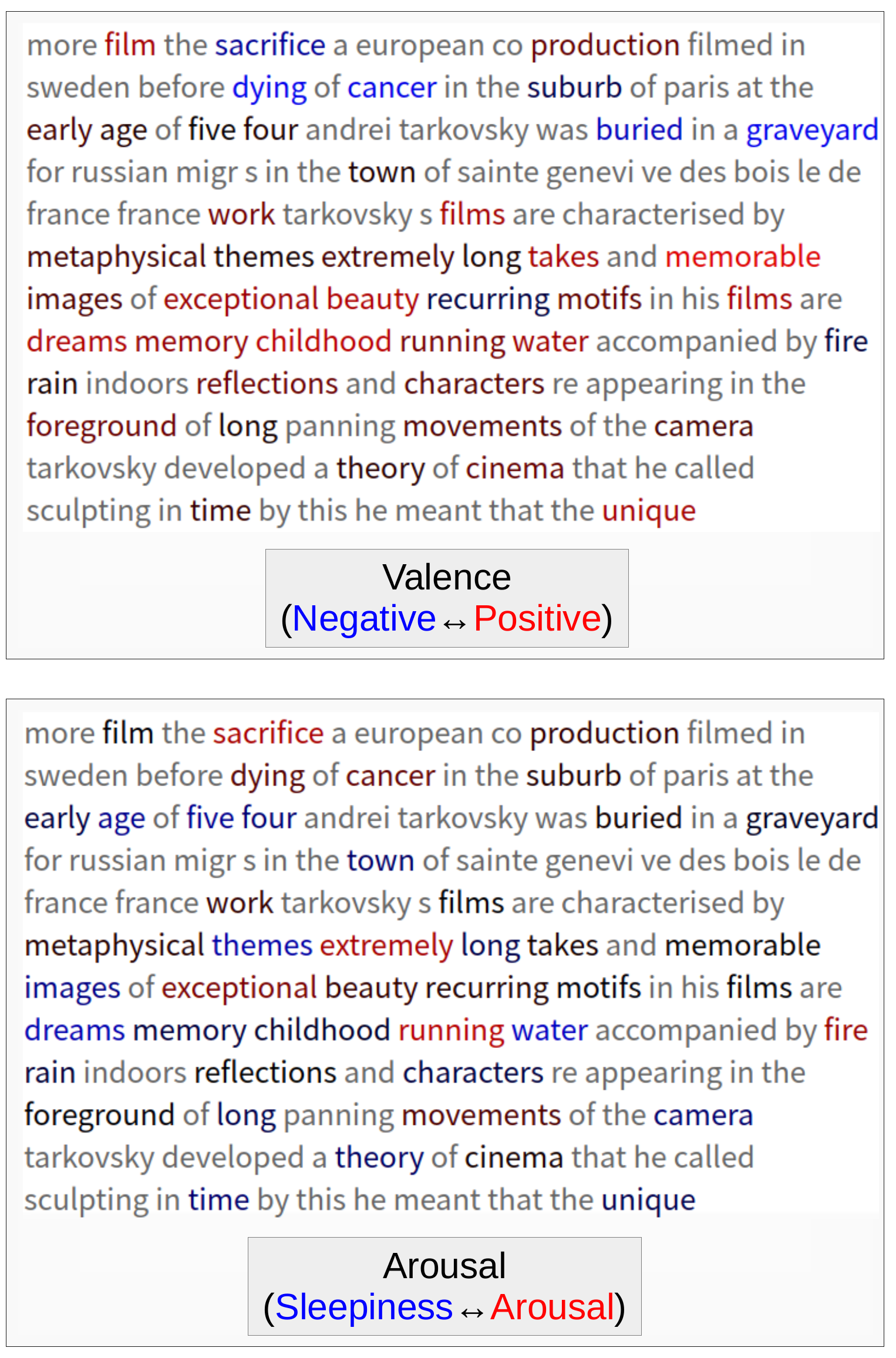}
		\caption{Distribution of valence and arousal levels in the text8 corpus} 
		\label{fig:visual}
	\end{center}
\end{figure}

\begin{figure*}[t]
	\begin{center}
		\includegraphics[width=12.5cm]{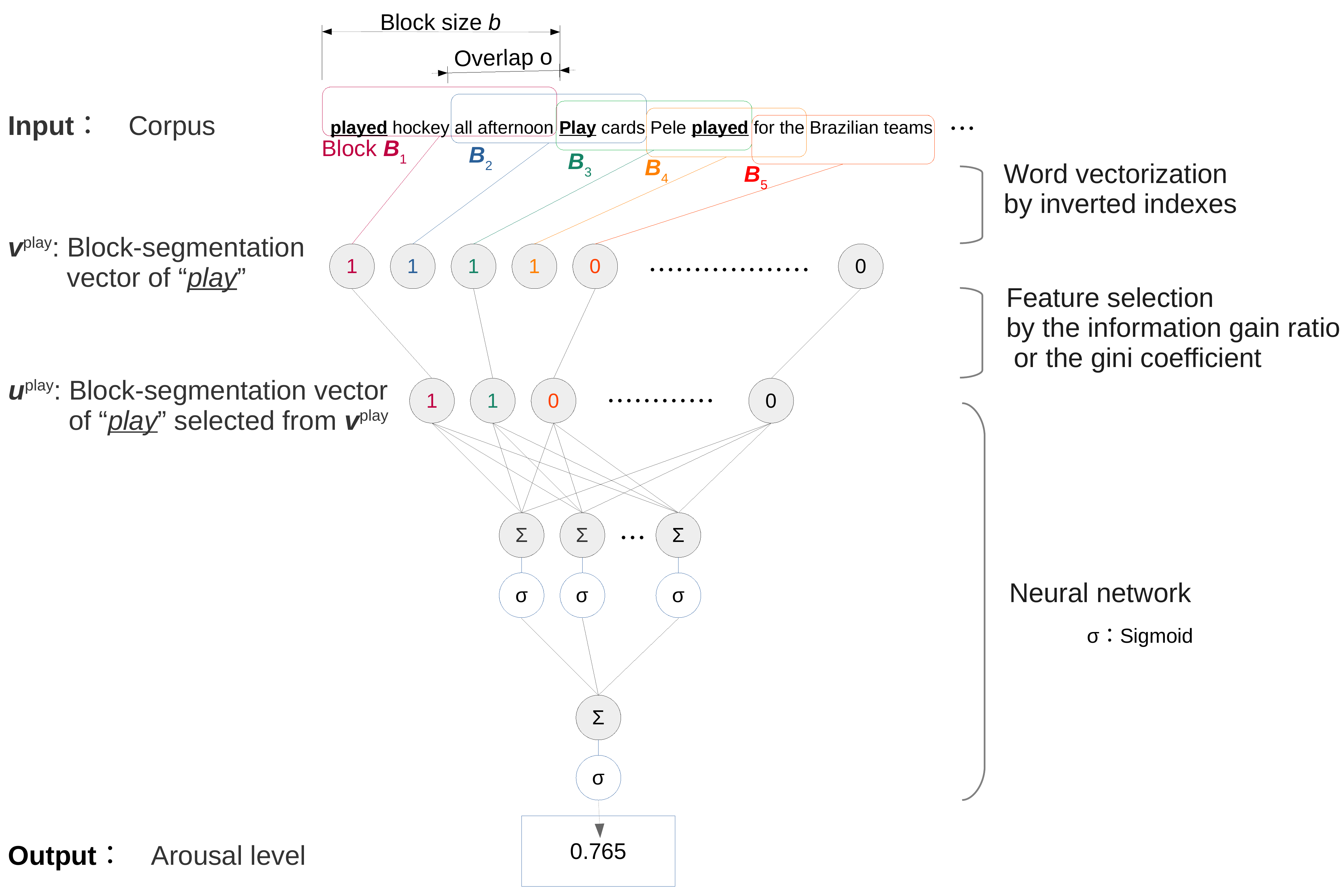}
		\caption{Architecture for arousal prediction using the word block-segmentation word vectors of a corpus} 
		\label{fig:BoBlockFlow}
	\end{center}
\end{figure*}

\subsection{p-normalized Kendall $\tau$ distance} 

In SentiWordNet, 1,105 synsets manually labeled in Micro-SentiWordNet are used as validation data and evaluated using Kendall's p-normalized rank correlation coefficient \cite{fagin2004comparing}. Let $S = \{ s_1, \ldots , s_h \}$ be the set of synsets in validation data. Let $o_{s_1}, \ldots , o_{s_h}$ be the ranks of predicted labels, and $\hat{o}_{s_1}, \ldots, \hat{o}_{s_h}$ be the ranks of labels in validation data, and let $p$ be the probability that a pair $(s_i, s_j)$ of synsets with the same rank $o_{s_i} = o_{s_j}$ in validation data is correctly ranked in the order of $\hat{o}_{s_1}, \ldots , \hat{o}_{s_j}$ for the predicted labels. Then, the evaluation value $\tau_p$ can be obtained as follows:
\begin{align}
S^2 &= \setdef{\{s_i,s_j\} }{ s_i, s_j \in S} \nonumber \\
n_{\textrm{d}} &= \left| \setdef{ \{s_i,s_j\} \in S^2 \!}{\! (o_{s_i} > o_{s_j} \land \hat{o}_{s_i} < \hat{o}_{s_j}) \lor (o_{s_i} < o_{s_j} \land \hat{o}_{s_i} > \hat{o}_{s_j}) } \right| \nonumber\\
n_{\textrm{u}} &= \left| \setdef{ \{s_i,s_j\} \in S^2 \!}{\! o_{s_i} = o_{s_j} \land \hat{o}_{s_i} \neq \hat{o}_{s_j} } \right| \nonumber\\
Z &= \frac{h(h-1)}{2} \nonumber\\
\tau_p &= \frac{n_{\textrm{d}} + p \cdot n_{\textrm{u}}}{Z}
\end{align}
where $0 \le \tau_p \le 1$. A low value of $\tau_p$ indicates that the predicted labels match the labels in validation data. In SentiWordNet, $p=\frac{1}{2}$ is used for evaluation.

\section{Prediction of Arousal Level} 

In this section, we present a learning method for arousal prediction using word vectors indexed by the block segmentation of a corpus.

\subsection{Distribution of value and arousal words in sentences} 

Russell's circumplex model \cite{russell1980circumplex} expresses emotions on the two axes: Positive$\leftrightarrow$Negative and Arousal$\leftrightarrow$Sleepiness. Figure \ref{fig:russell} plots major emotions in a circle on the axes of the emotion model.


We consider how positive and negative words as well as arousal and non-arousal words are distributed in samples of the text8 corpus. As shown in Figure \ref{fig:visual}, negative words are in blue, intermediate valence words are in black, and positive words are in red. By contrast, non-arousal is blue, intermediate arousal words are black, and arousal  words are red. In both samples, gray indicates words with no valence or arousal.


For valence, positive words such as "exceptional" and "beauty," "dreams" and "memory," and negative words such as "dying" and "cancer," "buried" and "graveyard," are clustered close together. In contrast, there is a mixture of arousal and non-arousal words in a row. For example, words with similar arousal levels, such as "dying" and "cancer," sometimes occur close together, but words with diﬀerent arousal levels, such as "themes" and "extremely" and "running" and "water," are sometimes mixed. These samples suggest that the valence of words can be discriminated using the surrounding words, but this is difficult to do for the arousal of words.


To capture the arousal levels, we need to identify the features of words with similar arousal levels in whole sentences. This is because only a part of the corpus, as shown in Figure 4, does not follow the occurrences of words with similar arousal levels together in other parts of the corpus.

\subsection{Block-segmentation word vector} 

Unlike words with similar valence, arousal words often do not appear near each other in a sentence. In SentiWordNet, glosses which are single sentences in WordNet cannot capture the flow of many sentences. Thus, we employ the text8 corpus of 17 million words extracted from Wikipedia sentences to capture a wide range of contexts. As the input of our learning method (Figure \ref{fig:BoBlockFlow}), we attempt to generate word vectors for supporting the behavior of arousal words in the context of sentences. The word vectors are devised to distinguish the contexts where arousal words are clustered together and words with different arousal levels are mixed together.


We express a corpus as a word sequence $T = \{w_1, \ldots , w_m\}$. Let $b$ be a block size and $o$ be an overlap with $0 \le o < 1$. For each block ID $i \in \{1, \ldots , n\}$, we define block $B_i$ in the corpus $T = B_1 \cup \cdots \cup B_n$ as follows:
\begin{align}
B_i = \{w_{(1-o) b i}, \ldots, w_{(1-o) b i + b}\}
\end{align}
Then, we define the block-segmentation vector $\bm{v}^w = \left( v^w_1, \ldots , v^w_n \right)$ of word $w$ on the inverted indexes for blocks $B_1 , \ldots , B_n$ as follows:
\begin{align}
v^w_i = 
\begin{cases}
1 & \textit{if } w \in B_i\\
0 & \textit{otherwise}
\end{cases}
\end{align}
Let $W_s$ be the set of synonymous words in synset $s$. Then, we define the block-segmentation vector $\bm{v}^s = \left(v^s_1, \ldots, v^s_n \right)$ of synset $s$ on the inverted indexes for blocks $B_1, \ldots , Bn$ as follows:
\begin{align}
\bm{v}^s &= \frac{1}{|W_s|} \sum_{w \in W_s} \bm{v}^{w}
\end{align}

\newcommand{\rg}{\bm{S}}
\newcommand{\vsi}{{v^s_i}}


From our perspective, the existing word embedding methods BoW and GloVe \cite{pennington2014glove} do not represent important features for arousal prediction in the distribution discussed in Section 3.1. Therefore, our word vectors are generated by a method in which the corpus is divided into blocks and an inverted index indicates the block ID containing each word. Although inverted indexes are often used for fast text searches \cite{zobel1998inverted}, they have not been widely used for word vectors in machine learning because of their high dimensionality.

\subsection{Feature selection}


By dividing a large corpus of sentences, the number $n$ of blocks in the corpus gives rise to a large number of dimensions of $\bm{v}^s$. We select some of the features in $\bm{v}^s$ based on the IGR and Gini coefficient for training data. Let $\rg$ be the set of synsets in the training data. Let $\rg^+$ be the set of synsets with arousal words and $\rg^-$ be the set of synsets with non-arousal words. Finally, let $v^s_i$ be the $i$th feature of vector $\bm{v}^s = (v_1^s, \ldots , v_n^s)$. We deﬁne the set $\rg^{\vsi}$ of synsets $s$ with $v^s_i \neq 0$ and the set $\rg^{-\vsi}$ of synsets $s$ with $v^s_i = 0$ as follows:
\begin{align}
\rg^{\vsi} = \{ s \in S \;|\; v^s_i \ne 0 \} \\
\rg^{-\vsi} = \{ s \in S \;|\; v^s_i = 0 \}
\end{align}
Then, the IGR for feature $v^s_i$ is defined as follows:
\begin{align}
\textit{Info}(\rg) &= - \sum_{c \in \{ +,- \}} \frac{|\rg^c|}{|\rg|} \log_2 \frac{|\rg^c|}{|\rg|} \nonumber \\
\textit{Info}_{\vsi}(\rg) &= - \sum_{x \in \{ \vsi,-\vsi \}} \frac{|\rg^x|}{|\rg|} \textit{Info}(\rg^x) \nonumber \\
\textit{SplitInfo}_{\vsi}(\rg) &= - \sum_{x \in \{ \vsi,-\vsi \}} \frac{|\rg^x|}{|\rg|} \log_2 \frac{|\rg^x|}{|\rg|} \nonumber \\
\textit{IG}_{\rg}(\vsi) &= \textit{Info}(\rg) - \textit{Info}_{\vsi}(\rg) \nonumber \\
\textit{IGR}_{\rg}(\vsi) &= \frac{\textit{IG}_{\rg}(\vsi)}{\textit{SplitInfo}_{\vsi}(\rg)}
\end{align}
Let $\epsilon_{\textrm{IGR}}$ be a border of IGR. We define the block-segmentation vector containing the features selected from $\bm{v}^s = \langle v^s_1 , \ldots , v^s_n \rangle $ by $\epsilon_{\textrm{IGR}}$ as follows:
\begin{align}
\bm{u}^s_{\epsilon_{\textrm{IGR}}} = \langle u^s_1, \ldots ,u^s_k \rangle
\end{align}
where $\{u^s_1, \ldots , u^s_k\} = \{ v \in \{v^s_1, \ldots , v^s_n\} \;|\; \Delta \textit{IGR}_{\rg}(v) \geq \epsilon_{\textrm{IGR}}\}$.

\newcommand{\gini}{\textit{Gini}}
Furthermore, the Gini coefficient for feature $\vsi$ is defined as follows:
\begin{align}
\gini(\rg) &= 1 -\sum_{c \in \{+,-\}} \left(\frac{|\rg^c|}{|\rg|}\right)^2 \nonumber \\
\gini _{\vsi}(\rg) &= \sum_{x \in \{ \vsi,-\vsi \}} \frac{|\rg^x|}{|\rg|} \gini(\rg^x) \nonumber \\
\Delta \gini(\vsi) &= \gini(\rg) - \gini _{\vsi}(\rg)
\end{align}
Let $\epsilon_{\textrm{Gini}}$ be a border of the Gini coefficient. We define the block-segmentation vector containing the features selected from $\bm{v}^s = \langle v^s_1 , \ldots , v^s_n \rangle $ by $\epsilon_{\textrm{Gini}}$ as follows:
\begin{align}
\bm{u}^s_{\epsilon_{\textrm{Gini}}} = \langle u^s_1, \ldots ,u^s_k \rangle
\end{align}
where $\{u^s_1, \ldots , u^s_k\} = \{ v \in \{v^s_1, \ldots , v^s_n\} \;|\; \Delta \gini (v) \geq \epsilon_{\textrm{Gini}} \textrm{ for } i \in \{1, \ldots , n\} \}$.

\subsection{Semi-supervised learning with neural networks} 


We use neural network (NN) models for the binary classiﬁcation and regression analysis of arousal levels. As in the semi-supervised learning steps of SentiWordNet, we obtain training data from synsets with a radius of $k$ labeled by following synonym links from seed data. The NRC-VAD-Lexicon is an emotional dictionary that labels each word with valence, arousal, and dominance levels. For the binary classiﬁcation of arousal, the synsets of words with arousal levels of $0.6$ or higher and $0.4$ or lower according to the NRC-VAD-Lexicon \cite{vadacl2018} are applied as positive and negative examples, respectively. For the regression analysis of arousal, the arousal levels of words in the NRC-VAD-Lexicon are represented by real numbers in $[0,1]$ and applied as numerical examples. The regression analysis leads to a learning model to distinguish the intensity of arousal in more detail. We predict an arousal label for each of the four radii $k\in\{0, 2, 4, 6\}$ to calculate the average of four predicted labels, where Arousal is 1 and Non-Arousal is 0.


\subsection{Parameters of the random walk step} 

To improve the predication accuracy in the random walk step, we define $\alpha$ in Equation (1), where $\alpha$ determines how much of the WordNet definition statement is reflected. After running the random walk from $0.0$ to $1.0$ in increments of $0.1$, the optimal values for validation data were found to be $\alpha = 0.9$ for valence and $\alpha = 0.7$ for arousal, as shown in Table \ref{table:randomwalk}. The optimal value of $\alpha$ for arousal is high but lower than that of valence, which indicates that arousal prediction is more difficult than valence prediction.

\begin{table}[t]
	\centering
	\caption{Value of $\tau_p$ for $\alpha$ in the random walk step}
	\label{table:randomwalk}
	\begin{tabular}[t]{c|ccc}
		\hline
		$\alpha$ & Arousal & Positive & Negative \\ \hline
		0.0 & 0.3444 & 0.2547 & 0.2291 \\
		0.1 & 0.3104 & 0.2280 & 0.2079 \\
		0.2 & 0.3101 & 0.2259 & 0.2077 \\
		0.3 & 0.3098 & 0.2239 & 0.2072 \\
		0.4 & 0.3095 & 0.2218 & 0.2066 \\
		0.5 & 0.3092 & 0.2199 & 0.2058 \\
		0.6 & 0.3090 & 0.2179 & 0.2051 \\
		0.7 & \textbf{0.3089} & 0.2167 & 0.2037 \\
		0.8 & 0.3092 & 0.2154 & 0.2027 \\
		0.9 & 0.3103 & \textbf{0.2146} & \textbf{0.2009} \\
		1.0 & 0.4885 & 0.2525 & 0.2793 \\ \hline
	\end{tabular}
\end{table}

\section{Experiments}

\setlength{\captionmargin}{10pt}

\begin{figure}[t]
	
	\centering
	\includegraphics[width=12.5cm]{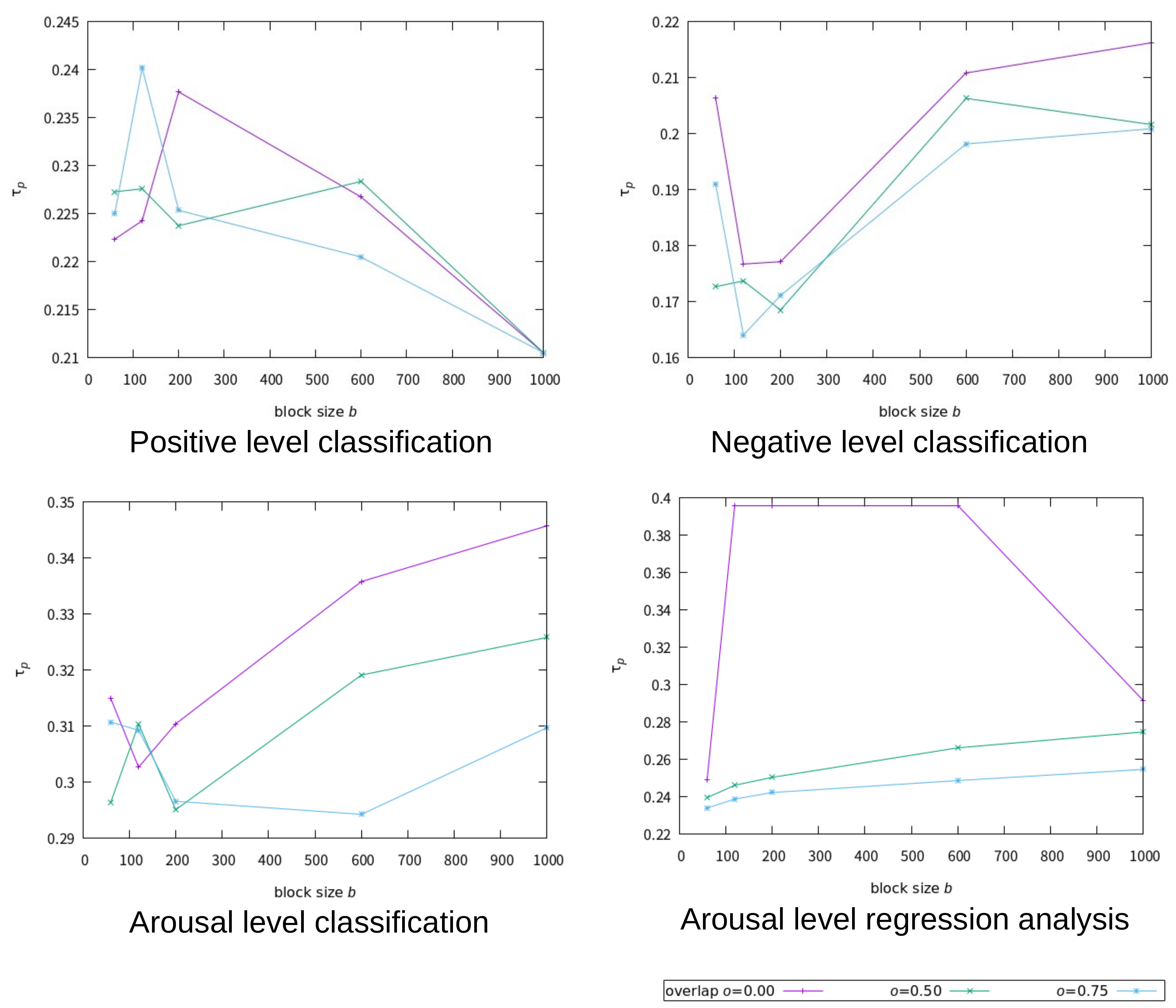}
	\caption{Results of valence and arousal levels using the block-segmentation vector with $\textit{IGR}_{\rg}(\vsi) \geq 0.01$}
	\label{fig:resultIGR}
			
\end{figure}

\begin{figure}[t]

	\centering
	\includegraphics[width=12.5cm]{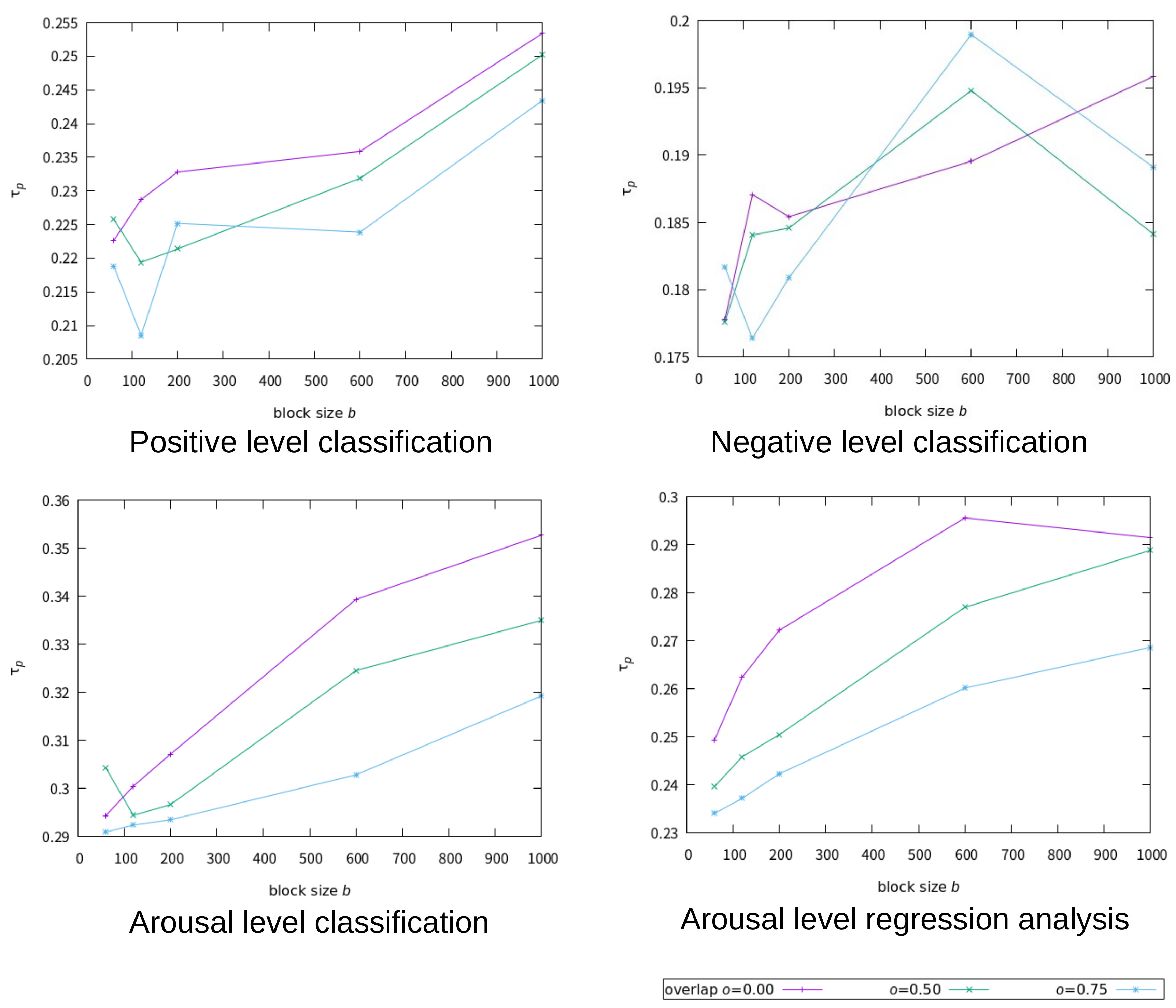}
	\caption{Results of valence and arousal levels using the block-segmentation vector with $\Delta \gini(\vsi) \geq \epsilon_{\textrm{Gini}}$}
	\label{fig:resultGini}

\end{figure}




We evaluate the results of valence and arousal prediction, which assigns levels to many WordNet synsets from a few labeled words. We evaluate the Kendall's p-normalized rank correlation coeﬀicient $\tau_p$ of the predicted labels where a lower value indicates a higher accuracy. In our experiments, we implemented the proposed arousal prediction method in Python and used a computer with an Intel Xeon W-2245 CPU running at 3.90GHz and 512 GB of memory.



For arousal prediction, we split target words in NRC-VAD-Lexicon, using 50\% as training data, 40\% as validation data, and 10\% as test data. The labeled words in the training data are only 10\% of all the synsets in WordNet. All words in the NRC-VAD-Lexicon are manually labeled with real numbers from 0 to 1 for each word's valence, arousal, and dominance. In our experiment, we used only the labels for arousal words. For valence prediction, we used the $\textit{Pwords}$ and $\textit{Nwords}$ as training data and the Micro-WNOp corpus \cite{cerini2007micro} as test data, based on the experimental setting of SentiWordNet.


For 256 epochs with the optimizer Adam and a batch size of 32, we trained an NN model with a sigmoid activation function on the input layer, hidden layer, and output layer by minimizing the cross entropy error loss function for binary classiﬁcation and the mean absolute error loss function for regression analysis. The hidden layer size is decreased at a constant rate, and the output layer size is two.

\subsection{Results for block-segmentation word vectors} 

Figure \ref{fig:resultIGR} shows the accuracy of valence and arousal prediction using the block-segmenta-\\tion vector for each block size $b \in \{60, 120, 200, 600, 1000\}$  and each overlap $o \in \{0.00, 0.50, 0.75\}$ with an IGR border of $\epsilon_{\textrm{IGR}}=0.01$. For a larger overlap, the value of $\tau_p$ is improved by the proposed word vectors for supporting overlaps. The results with overlap $o=0.75$ are better than those with $o=0.50$ and $o=0.00$, indicating that the appropriate overlap depends on the block size. Moreover, a smaller block size is better because extremely large block sizes, such as 600 or 1,000, reduce the accuracy. Namely, if block size $b$ is reduced and overlap $o$ is increased, the accuracy is improved, but the dimensionality of the word vectors is increased, resulting high computation time and memory costs. The dimensionality can be reduced by a low IGR border $\epsilon_{\textrm{IGR}}$, but the accuracy will decrease.



Figure \ref{fig:resultGini} also shows the accuracy of valence and arousal prediction using the block segmentation vector for each block sizes $b \in \{60, 120, 200, 600, 1000\}$ and each overlap $o \in \{0.00, 0.50, 0.75\}$ with the Gini coefficient. The border of the Gini coefficient $\epsilon_{\textrm{Gini}}$ is given by the median of $\Delta \gini(f_i)$ in all $i \in \{1, \ldots, n\}$. The results with the Gini coeﬃcient are improved for small block sizes when compared with the results of feature selection using the IGR.

\subsection{Comparison with existing methods in SentiWordNet} 

Table \ref{table:result} summarizes the $\tau_p$ values of valence and arousal levels predicted by our methods and previous methods using support vector machines and NNs.
For block segmentation vectors, we chose the block sizes $b \in \{1320, 1440, 1560, 1680, 1800, 1920, 2040, 2160, \\2280, 2400\}$ and the overlap $o \in \{0.00, 0.50, 0.75\}$ based on the validation set performance. 
The results of SVM-BoW in SentiWordNet are $\tau_p = 0.281$ for Positive$\leftrightarrow$Not-Positive and $\tau_p = 0.231$ for Negative$\leftrightarrow$Not-Negative. NN-BoW in SentiWordNet improves some results ($\tau_p = 0.215$ for Positive$\leftrightarrow$Not-Positive and $\tau_p = 0.202$ for Negative$\leftrightarrow$Not-Negative), but worsens others ($\tau_p = 0.392$ for the binary classiﬁcation of Arousal $\leftrightarrow$ Non-Arousal and $\tau_p = 0.361$ for the regression analysis of Arousal$\leftrightarrow$Non-Arousal). The regression analysis result is better than the binary classiﬁcation result, but not enough when compared with $\tau_p = 0.281$ for Positive$\leftrightarrow$Not- Positive in SVM-BoW, indicating the diffculty of predicting arousal.

Note that the short sentences in the WordNet glosses fail to characterize the features of words for arousal prediction. We used the text8 corpus on Wikipedia which contains long sentences in articles to generate word vectors from the distribution of arousal words and non-arousal words.

To compare our block-segmentation vector (BSeg) with existing methods, we show the results of GloVe for predicting valence and arousal levels. The results of NN-GloVe in SentiWordNet are $\tau_p = 0.207$ for Positive$\leftrightarrow$Not-Positive, $\tau_p = 0.151$ for Negative$\leftrightarrow$Not-Negative, $\tau_p$ = 0.328 for the binary classiﬁcation of Arousal$\leftrightarrow$Non-Arousal, and $\tau_p = 0.300$ for the regression analysis of Arousal$\leftrightarrow$Non-Arousal. All the results are better than those of SVM-BoW and NN-BoW in SentiWordNet, but further improvement is needed for arousal prediction.

For arousal prediction, our methods NN-BSeg-IGR and NN-BSeg-Gini outperform the results of the existing methods SVM-BoW, NN-BoW, and NN-GloVe in SentiWordNet. 
In particular, NN-BSeg-Gini with a block size of $b = 2400$ and an overlap of $o = 0.75$ achieves the best value $\tau_p = 0.217$ for the regression analysis of Arousal$\leftrightarrow$Non-Arousal. The features of word vectors were decrased from 113,384 to 28,846 by the IGR and to 23,029 by the Gini coefficient. However, the $\tau_p$ values of our methods ($\tau_p = 0.232$ and $0.174$) for valence prediction are worse than the $\tau_p$ values of NN-GloVe ($\tau_p = 0.207$ and $0.151$). That is, GloVe is effective for predicting valence but not arousal because arousal is orthogonal to valence. For both valence and arousal prediction, the average $0.208$ of the $\tau_p$ values obtained by the proposed method outperforms the average $\tau_p$ values of all the existing methods SVM-NN, NN-BoW, and NN-GloVe. Therefore, the block segmentation vectors in NN-BSeg-IGR and NN-BSeg-Gini successfully represent important features for predicting arousal levels.





\section{Conclusions}



We proposed a block-segmentation vector for predicting the arousal levels of Word- Net synsets using semi-supervised learning, which is more difficult than predicting the valence levels. We analyzed the diﬀerent distributions of valence and arousal words, and demonstrated that the previous methods in SentiWordNet are not suitable for arousal prediction. Based on this analysis, we divided the Wikipedia corpus into blocks to identify the occurrences of arousal and non-arousal words in the context of long sentences. We introduced block-segmentation vectors based on inverted indexes with the block IDs. The experimental results show that our block-segmentation vectors with feature selection improve the accuracy of arousal prediction when compared with the existing methods of BoW and GloVe in SentiWordNet.

Our future work will be to expand the block-segmentation vectors using part-of- speech information to improve sentiment analysis according to the content of sentences. Furthermore, we will implement an emotion analysis system by combining other emotion perspectives and models with those for valence and arousal.

\begin{table*}[t]
	\caption{Valence and arousal $\tau_p$ values using conventional methods with BoW and GloVe and our method with block-segmentation vectors} 
	\label{table:result}
	\centering
	\scalebox{0.85}{
		\begin{tabular}[t]{|c|c|c||ccc|c|}
			\hline
			\multicolumn{2}{|c|}{}&& Positive & Negative & Arousal  & \\
			\multicolumn{2}{|c|}{learning method} & learning task& $\updownarrow$ & $\updownarrow$ & $\updownarrow$ & Average \\ 
			\multicolumn{2}{|c|}{} && Non- & Non- & Non- & \\
			\multicolumn{2}{|c|}{} && Positive & Negative & Arousal & \\ \hline \hline
			& SVM-BoW& Classification & 0.281 & 0.231 & 0.414 & 0.309 \\ \cline{2-7}
			SentiWordNet & NN-BoW &  $\begin{array}{c} \textrm{Classification} \\ \textrm{Regression} \end{array}$ & $\begin{array}{c} \textrm{0.215} \\ \textrm{ } \end{array}$ & $\begin{array}{c} \textrm{0.202} \\ \textrm{ } \end{array}$ & $\begin{array}{c} 0.392 \\ 0.361 \end{array}$ &  $\begin{array}{c} 0.270 \\ 0.259 \end{array}$ \\ \cline{2-7}
			& NN-GloVe &  $\begin{array}{c} \textrm{Classification} \\ \textrm{Regression} \end{array}$ & $\begin{array}{c} \textrm{\textbf{0.207}} \\ \textrm{ } \end{array}$ & $\begin{array}{c} \textrm{\textbf{0.151}} \\ \textrm{ } \end{array}$ & $\begin{array}{c} 0.328 \\ 0.300 \end{array}$ & $\begin{array}{c} \textbf{0.229} \\ 0.219 \end{array}$ \\ \hline
			our method & NN-BSeg-IGR & $\begin{array}{c} \textrm{Classification} \\ \textrm{Regression} \end{array}$ & $\begin{array}{c} \textrm{0.225} \\ \textrm{ } \end{array}$ & $\begin{array}{c} \textrm{0.191} \\ \textrm{ } \end{array}$ & $\begin{array}{c} 0.331 \\ 0.234 \end{array}$ & $\begin{array}{c} 0.242 \\ 0.217 \end{array}$ \\ \cline{2-7}
			& NN-BSeg-Gini &  $\begin{array}{c} \textrm{Classification} \\ \textrm{Regression} \end{array}$ & $\begin{array}{c} \textrm{0.232} \\ \textrm{ } \end{array}$ & $\begin{array}{c} \textrm{0.174} \\ \textrm{ } \end{array}$ & $\begin{array}{c} \textbf{0.281} \\ \textbf{0.217} \end{array}$ & $\begin{array}{c} \textbf{0.229} \\ \textbf{0.208} \end{array}$ \\
			\hline
		\end{tabular}
	}
\end{table*}

\addcontentsline{toc}{section}{References}
\bibliography{bibfile}
\bibliographystyle{tieice.bst}

\end{document}